\documentclass[letterpaper, 10 pt, conference]{IEEEtran}

\usepackage{amsmath}
\usepackage{amssymb}
\usepackage{hyperref}
\usepackage{xcolor}
\usepackage{tikz}
\usepackage{longtable}
\usepackage{booktabs}
\usepackage{comment}
\usepackage{nccmath}
\usetikzlibrary{positioning, matrix, fit, calc, arrows.meta, patterns, decorations.pathreplacing, calligraphy}

\newcommand{\CN}[1]{\textcolor{red}{CN: #1}}
\newcommand{\LW}[1]{\textcolor{orange}{LW: #1}}
\newcommand{\TK}[1]{\textcolor{blue}{TK: #1}}
\newcommand{\LP}[1]{\textcolor{green}{LP: #1}}
\newcommand{\EB}[1]{\textcolor{purple}{EB: #1}}
\newcommand{\ALL}[1]{\textcolor{magenta}{ALL: #1}}

\renewcommand{\CN}[1]{}
\renewcommand{\LW}[1]{}
\renewcommand{\TK}[1]{}
\renewcommand{\LP}[1]{}
\renewcommand{\EB}[1]{}
\renewcommand{\ALL}[1]{}

\newtheorem{definition}{Definition}

\title{On Quantification for SOTIF Validation of Automated Driving Systems}

\author{
	\IEEEauthorblockN{Lina Putze, Lukas Westhofen, Tjark Koopmann, Eckard B\"ode and Christian Neurohr}
	\IEEEauthorblockA{German Aerospace Center (DLR) e.V.\\
		Institute of Systems Engineering for Future Mobility, Oldenburg, Germany\\
		Email: firstname.lastname@dlr.de}
}

\begin{document}

\maketitle

\begin{abstract}
Automated driving systems are safety-critical cyber-physical systems whose safety of the intended functionality (SOTIF) can not be assumed without proper argumentation based on appropriate evidences. 
Recent advances in standards and regulations on the safety of driving automation are therefore intensely concerned with demonstrating that the intended functionality of these systems does not introduce unreasonable risks to stakeholders. 
In this work, we critically analyze the ISO 21448 standard which contains requirements and guidance on how the SOTIF can be provably validated. 
Emphasis lies on developing a consistent terminology as a basis for the subsequent definition of a validation strategy when using quantitative acceptance criteria. 
In the broad picture, we aim to achieve a well-defined risk decomposition that enables rigorous, quantitative validation approaches for the SOTIF of automated driving systems.
\end{abstract}

\section{Introduction}
\label{sec:introduction}

Ensuring safe operation of automated driving systems (ADSs) is paramount for their broad introduction to public traffic.
Recently, the UN regulation no. 157 was introduced \cite{unece157}, containing requirements on the type approval of automated lane keeping systems, an ADS according to SAE J3016 \cite{sae2021definitions}.
In fact, the Mercedes-Benz Drive Pilot is an ADS that has been granted type approval for German roads in 2022 while meeting the requirements of the UN regulation no. 157, according to the manufacturer.
This regulation is also adopted, for example, in the German act on the approval and operation of motor vehicles with autonomous driving functions (AFGBV) \cite{germanadsact}.

One aspect of such a type approval is demonstrating the system's safety. 
For this, the industry has developed standards such as ISO 26262 \cite{iso26262} and ISO 21448 \cite{iso21448}.
Whereas the first is concerned with functional safety, i.e.\ risks arising from non-compliance of items to their specification, the ISO 21448 is concerned with safety of the intended functionality (SOTIF), i.e.\ with risks arising from the specification itself.
Specifically, one requirement of the UN regulation no. 157 reads: 
'[Auditors/assessors] shall in particular be competent as auditor/assessor for [\dots] ISO 21448' \cite{unece157}.
For Germany, the AFGBV additionally requires assessing the system safety compliant with the state of the art \cite[Annex 7.2]{germanadsact}.
According to the AFGBV, compliance with the ISO 21448 is a \emph{sufficient} criterion for a state of the art safety assessment. 
Thus, both manufactures and type approval authorities must consider SOTIF as a cornerstone.

In order to achieve SOTIF, issues of the completeness and safety of the specification in all possible operational scenarios need to be addressed. 
In general, SOTIF gained traction due to increasing levels of driving automation and their interactions with ordinary traffic participants in open contexts, e.g.\ because adversarial environments not being correctly considered when specifying functionality. 
Due to this, the ISO 21448 was initiated in 2016, publicly available as a specification since 2019, and finally released in June 2022.

As such, ISO 21448 compliant safety cases can now be conducted, which is, as stated above, essential for compliance with the UN regulation no. 157.
Moreover, the safety case's rigor benefits drastically by an at least partially quantitative validation.
Hence, our research question is: \emph{how does the ISO 21448 require or suggest performing the quantitative SOTIF validation}?
In the end, it must be possible to devise a validation strategy implementing the standard's requirements.

This necessitates a rigorous investigation of the ISO 21448.
To our knowledge, this work is the first to perform such an investigation as its main topic. 
In order to fully understand how the standard handles risk-related quantities, acceptance criteria and validation targets, we analyze in detail the respective definitions, their relations as well as the normative and informative SOTIF validation aspects within this standard. 
Specifically, we
\begin{enumerate}
	\item study the terminological risk framework as well as the relevant normative and informative parts on SOTIF validation within the ISO 21448 in depth,
	\item critically debate where the content is insufficient for the implementation of a compliant, quantitative SOTIF validation method, and  
	\item provide constructive suggestions for its improvement.
\end{enumerate}

Traditionally, we discuss related work in \autoref{sec:related_work}.
We continue with examining and debating the terminological risk framework in \autoref{sec:relation_of_definitions} and aspects of quantitative validation in \autoref{sec:quantification_concept_phase}.
The endeavor concludes with \autoref{sec:conclusion}.

\section{Related Work}
\label{sec:related_work}

Historically, assessment of active safety characteristics in the automotive industry considered mainly functional safety for manually driven vehicles, i.e.\ risks arising from item malfunctions.
For this, the ISO 26262 is the predominant source \cite{iso26262}.
It defines a terminological risk framework around exposure, controllability and severity.
These components can be assessed quantitatively and then categorized into discrete levels, cf.~e.g.\ Krampe and Junge for severity \cite{krampe2020injury}.

When moving towards higher automation levels, risks arising from the specification itself become more relevant -- and so does the need for rigorous quantification of associated safety aspects.
For this, there exist approaches instantiating the requirements posed by the ISO 21448 \cite{iso21448}.
Prior work, conducted within PEGASUS (\href{https://www.pegasusprojekt.de/en/}{www.pegasusprojekt.de/en}), seeks to quantify exposure by estimating the probability of occurrence of the triggering environmental conditions \cite{kramer2020identification}. 

A comprehensive survey of risk assessment approaches for automated driving has been conducted by Chia et al. \cite{chia2022risk}. 
Therein, the authors classify risk assessment methods along to their properties, including whether they are qualitative or quantitative, and if they address functional safety, SOTIF or both.
In contrast to the aforementioned survey paper, this work aims to mature the ISO 21448 itself for such quantitative methods. 
The closest related work by Zhu et al., who propose a systematic identification of SOTIF triggering conditions as their main contribution, already discovered some of these inconsistencies and subsequently used their own adapted definitions \cite{zhijing2022}. 
Let us also mention the work of Saberi et al., who challenged the ISO 26262/21448 standards' sufficiency regarding emergent behavior resulting from unintended functionality while the ISO 21448 was still under development \cite{sabari2020}.

Concerning risk quantification for ADSs, de Gelder et al.\ suggest a particularly relevant approach \cite{de2021risk}. 
They model the risk associated with a scenario class as a product of the expected values of probability distributions of exposure, severity and controllability, following the ISO 26262.
In their case study, the exposure to scenario classes is estimated from real-world data that resulted from letting experienced human drivers follow a prescribed route, while the expected values for the severity and controllability are estimated using simulations. 
Regarding the ISO 21448, they incorporated two triggering conditions in their simulations, however, without considering their probability of occurrence. 
Even though the exposure to scenario classes can not be independent of the ADS and their modeling of controllability as well as the simulation validity are debatable, the work at hand was inspired by the idea of a rigorous, quantitative risk modeling for the SOTIF.

Another approach by Buerkele et al.\ proposes to estimate the probability of harm based on mutual influences of perception failures and hazards induced by the environment \cite{buerkle2022modelling}. 

Obviously, risk quantification is also examined in other domains. 
On a meta level, the ISO/IEC Guide 51 gives (terminological) recommendations for standard developers \cite{iso51}.
For domain-independent system safety, a well-known hazard analysis framework is given by Ericson \cite{ericson2015hazard}.
For the railway and aviation domains, Filip et al.\ attempt to adopt safety quantification frameworks to ADSs \cite{filip2022derivation}.
For the civil aerospace domain, comprehensive risk assessment procedures exist, addressing both functional safety and topics comparable to SOTIF. 
The overall safety process is defined by ARP4754A \cite{arp4754a} 
and complemented by ARP4761 \cite{arp4761}. 
The latter provides specific guidance on how to execute the safety analysis, generate the necessary evidences and assemble an overall safety argumentation for failure conditions (FCs) on aircraft level. 
Such top-level FCs are classified into five severity categories: \emph{no} safety impact, \emph{minor}, \emph{major}, \emph{hazardous} and \emph{catastrophic}. 
Quantitative safety targets are defined for FCs \emph{major} and above (e.g.\ for \emph{catastrophic} FCs failure rates must be $< 10^{-9}/h$).

\section{Terminological Framework of the ISO 21448}
\label{sec:relation_of_definitions}

\begin{figure*}
	\centering
	\scalebox{0.9}{

\begin{tikzpicture}[
		every node/.append style={draw, text width=1.6cm, font=\scriptsize, align=center, minimum height=2em, inner sep=1pt},
		21448defined/.style={},
		21448and26262undefined/.style={fill=gray!20},
		21448adoptsfrom26262/.style={pattern=crosshatch dots,pattern color=gray!50},
		26262defined21448undedfined/.style={pattern=north west lines, pattern color=gray!50},
		edge/.style={-{Stealth[length=1.5mm]}}
	]
	

	\node[21448adoptsfrom26262]											(reasonably-foreseeable)									{Reasonably Foreseeable};
	\node[21448defined, right=0.9cm of reasonably-foreseeable]			(misuse)													{Misuse};
	\node[21448and26262undefined, right=0.9cm of misuse]				(hazardous-behavior)										{Hazardous Behavior};
	\node[21448defined, right=0.9cm of hazardous-behavior]				(scene)														{Scene};
	\node[21448and26262undefined, right=0.9cm of scene]					(situation)													{Situation};
	\node[21448defined, right=0.9cm of situation]						(event)														{Event};
	\node[21448adoptsfrom26262, right=0.9cm of event]					(harm)														{Harm};

	\node[21448defined, below=0.3cm of scene]							(action)													{Action};
	\node[26262defined21448undedfined, below=0.3cm of event]			(hazard)													{Hazard};
	\node[21448adoptsfrom26262, below=0.3cm of harm]					(controllability)											{Controllability};
	
	\node[21448defined, below=2*0.3cm+2em of situation]					(scenario)													{Scenario};
	
	\node[21448defined, below=2*0.3cm+2em of action]					(triggering-condition)										{Triggering Condition};
	\node[21448adoptsfrom26262, below=0.3cm of scenario]				(operational-situation)										{Operational Situation};
	
	\node[21448defined, below=4*0.3cm+3*2em of misuse]					(performance-insufficiency)									{Performance Insufficiency};
	\node[21448defined, below=4*0.3cm+3*2em of hazardous-behavior]		(insufficiency-of-specification)							{Insufficiency of Specification};
	\node[21448and26262undefined, below=0.3cm of triggering-condition]	(occurrence)												{Occurrence};
	\node[21448adoptsfrom26262, below=0.3cm of operational-situation]	(exposure)													{Exposure};
	\node[21448adoptsfrom26262, below=3*0.3cm+2*2em of hazard]			(hazardous-event)											{Hazardous Event};
	
	\node[21448defined, below=0.3cm of insufficiency-of-specification]	(functional-insufficiency)									{Functional Insufficiency};
	\node[21448adoptsfrom26262, below=0.3cm of hazardous-event]			(severity)													{Severity};
	\node[21448defined, below=4*0.3cm+3*2em of controllability]			(validation-target)											{Validation Target};
	
	\node[26262defined21448undedfined, below=0.3cm of functional-insufficiency]		(intended-functionality)									{Intended Functionality};
	\node[26262defined21448undedfined, below=0.3cm of severity]			(risk)														{Risk};
	\node[21448defined]													(sotif) at ($(occurrence |- risk)!0.5!(exposure |- risk)$)	{\bfseries SOTIF};
	\node[21448defined, below=0.3cm of validation-target]				(acceptance-criterion)										{Acceptance Criterion};
	
	
	\draw[edge] ([xshift=-0.3cm]reasonably-foreseeable.south) |- (performance-insufficiency);
	\draw[edge] (reasonably-foreseeable) |- ([xshift=0.2cm,yshift=0.05cm] insufficiency-of-specification |- triggering-condition) -- ([xshift=0.2cm]insufficiency-of-specification.north);
	\draw[edge] ([xshift=0.3cm]reasonably-foreseeable.south) |- ([yshift=0.15cm]triggering-condition.west);
	\draw[edge] ([xshift=-0.3cm]misuse.south) -- ([xshift=-0.3cm]performance-insufficiency.north);
	\draw[edge] (misuse) -- ($(misuse)!0.77!(performance-insufficiency)$) -| ([xshift=-0.2cm]insufficiency-of-specification.north);
	\draw[edge] ([xshift=0.3cm]misuse.south) |- ([yshift=0.25cm]triggering-condition.west);	
	\draw[edge] ([xshift=-0.3cm]hazardous-behavior.south) -- ([xshift=-0.3cm]$(hazardous-behavior)!0.7!(hazardous-behavior |- triggering-condition)$) -| ([xshift=0.5cm]performance-insufficiency.north);
	\draw[edge] (hazardous-behavior) -- (insufficiency-of-specification);
	\draw[edge] ([xshift=0.3cm]hazardous-behavior.south) -- ([xshift=0.3cm]$(hazardous-behavior)!0.7!(hazardous-behavior |- triggering-condition)$) -| ([xshift=-0.3cm]triggering-condition.north);
	\draw[edge] (hazardous-behavior.east) -| ([yshift=0.55cm]$(hazard)!0.84!(hazardous-behavior |- hazard)$) -- ([xshift=-0.7cm,yshift=0.55cm]hazard.west) |- ([yshift=0.2cm]hazard.west);
	\draw[edge] (scene) -- (action);
	\draw[edge] (scene.east) -| ([yshift=0.3cm]$(scenario)!0.5!(scene |- scenario)$) -- ([yshift=0.3cm]scenario.west);
	\draw[edge] (situation) -- (scenario);
	\draw[edge] (event.west) -| ($(scenario)!0.5!(hazard |- scenario)$) -- (scenario.east);
	\draw[edge] ([yshift=0.2cm]harm.west) -| ([xshift=-0.2cm]$(hazard)!0.5!(controllability)$) -- (hazard.east);
	\draw[edge] (harm.west) -| ($(severity)!0.5!(validation-target)$) -- (severity.east);
	\draw[edge] ([yshift=-0.2cm]harm.west) -| ([xshift=0.2cm,yshift=0.2cm]$(risk)!0.5!(acceptance-criterion)$) -- ([yshift=0.2cm]risk.east);
	\draw[edge] (harm) -- (controllability);
	
	\draw[edge] (action) |- ([yshift=0.3cm]scenario);
	\draw[edge] (hazard) -- (hazardous-event);
	\draw[edge] (hazard.west) -| ([yshift=0.2cm]$(sotif)!0.72!(hazard |- sotif)$) -- ([yshift=0.2cm]sotif.east);
	
	\draw[edge] ([yshift=-0.3cm]scenario) -| (triggering-condition);
	\draw[edge] (scenario) -- (operational-situation);
	
	\draw[edge] ([yshift=-0.18cm]triggering-condition.west) -| (performance-insufficiency);
	\draw[edge] ([yshift=-0.3cm]triggering-condition.west) -| ($(insufficiency-of-specification)!0.5!(occurrence)$) -- (insufficiency-of-specification.east);
	\draw[edge] (operational-situation) -- (exposure);
	\draw[edge] (operational-situation) -| ([xshift=-0.6cm]hazardous-event);
	
	\draw[edge] (performance-insufficiency) |- (functional-insufficiency);
	\draw[edge] (insufficiency-of-specification) -- (functional-insufficiency);
	\draw[edge] (hazardous-event) -- (severity);
	
	\draw[edge] (functional-insufficiency) -| (sotif);
	\draw[edge] (severity) -- (risk);
	
	\draw[edge] (risk) -- (acceptance-criterion);
	\draw[edge] (risk) -- (sotif);
	\draw[edge] (intended-functionality) -- (sotif);
	
	\draw[edge] (acceptance-criterion) -- (validation-target);
	
	\matrix(key)[matrix of nodes, draw=none, nodes={anchor=west}, row sep=0mm, below left=0.15cm and -0.65cm of performance-insufficiency,font=\scriptsize]{
		\node[minimum height=1em, text width=0.3cm, 21448defined] {$T$}; & \node[align=left, draw=none, minimum height=1.1em, inner sep=3pt, text width=4.4cm] {\setlength{\baselineskip}{6pt}$T$ exclusively defined in ISO 21448\par}; \\
		\node[minimum height=1em, text width=0.3cm, 26262defined21448undedfined] {$T$}; & \node[align=left, draw=none, minimum height=1.1em, inner sep=3pt, text width=4.4cm] {\setlength{\baselineskip}{6pt}$T$ defined in ISO 26262, explicitly adopted in ISO 21448\par}; \\
		\node[minimum height=1em, text width=0.3cm, 21448adoptsfrom26262] {$T$}; & \node[align=left, draw=none, minimum height=1.1em, inner sep=3pt, text width=4.4cm] {\setlength{\baselineskip}{6pt}$T$ defined in ISO 26262, not explicitly referred to in ISO 21448\par}; \\
		\node[minimum height=1em, text width=0.3cm, 21448and26262undefined] {$T$}; & \node[align=left, draw=none, minimum height=1.1em, inner sep=3pt, text width=4.4cm]  {\setlength{\baselineskip}{6pt}$T$ defined neither in ISO 21448 nor in ISO 26262\par}; \\
	};
	\node[inner sep=1pt, dashed, fit=(key)] {};
	\node[above right=0cm and 0cm of key.north west, draw=none, inner sep=0, minimum height=1em, text width=0.5cm] {\emph{Key}};
	
\end{tikzpicture}}
	\caption{Relations of definitions in the ISO 21448 relevant for risk evaluation. An edge from terms $A$ to $B$ indicates that $A$ is used in $B$'s definition.}
	\label{fig:relation_definitions_21448}
\end{figure*}
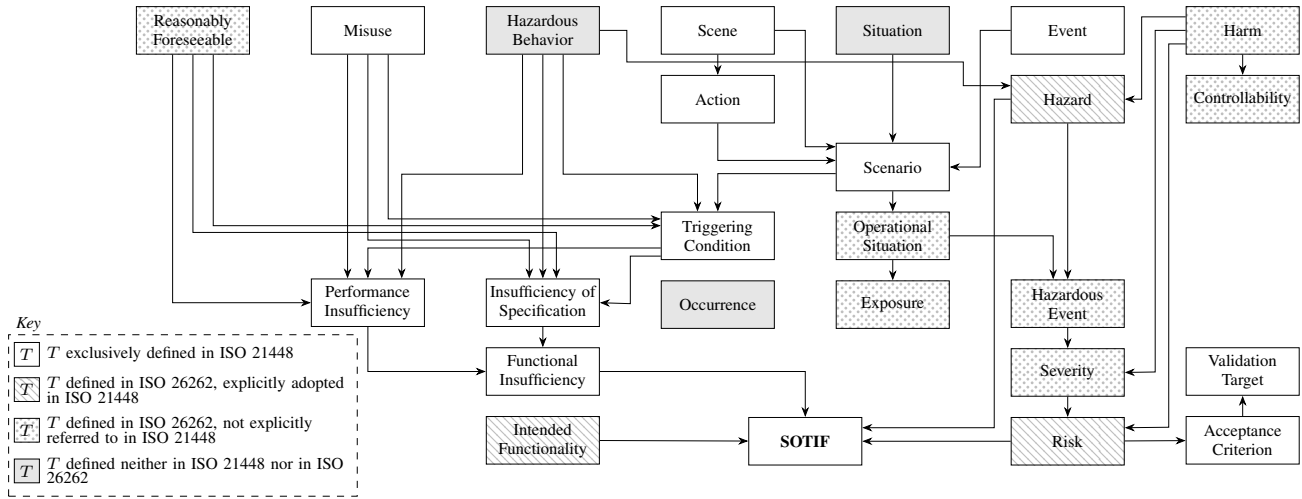

In order to grasp the role of quantification within the ISO 21448's argumentation framework it is paramount to pervade the underlying terms and their usage in the standard. 
Principally, the ISO 21448's terms and definitions are adopted from ISO 26262-1 and complemented by additions introduced in its Clause 3. 
\autoref{fig:relation_definitions_21448} depicts how the definitions of terms relevant to quantification, as suggested by the ISO 21448, build on each other as well as the terms' origin.

\subsection{Factual Analysis}\label{subsec:factual_analysis_terminology}

The ISO 21448's main goal is to ensure the \textit{SOTIF}, defined as the 'absence of unreasonable risk due to [some specific] hazards' \cite[3.25]{iso21448}. 
Therefore, in order understand this intention, comprehension of how the terms \textit{risk} and \textit{hazard} are applied within the ISO 21448 standard is essential.

The term \textit{risk} is already defined in the ISO 26262. 
Its definition as  'combination of the probability of occurrence of harm and the severity of that harm' \cite[1-3.128]{iso26262} is adopted by the ISO 21448. 
It builds on the term \textit{harm}  defined by the ISO 26262 as 'physical injury or damage to the health of persons' \cite[Part 1, 3.74]{iso26262}. 
Let us remark that the definition of \textit{harm} given by the ISO 26262 refers solely to the damage to persons and excludes damage to property in contrast to other standards like the ISO/IEC Guide 51 \cite{iso51}.
Besides \textit{harm}, \textit{risk} is based on the term \textit{severity}, which
is defined as an 'estimate of the extent of harm to one or more individuals that can occur in a potentially hazardous event ' \cite[1-3.154]{iso26262}. Hence, \textit{severity} builds not only on \textit{harm} but also on \textit{hazardous event}.

The term \textit{hazardous event} comes along with ambiguities, as
for its interpretation the ISO 21448's reader has two options: 
\begin{enumerate}
	\item[i)] refer to the definition of \textit{event} as an 'occurrence at a point in time' \cite[3.7]{iso21448} given by the ISO 21448, or
	\item[ii)] apply the ISO 26262's definition of \textit{hazardous event} as 'combination of a hazard and an operational situation'.
\end{enumerate}
In case i), the meaning of the adjective \textit{hazardous} is unclear.
The  ISO 21448's Figures~1, 4, 12 and 13 depict the \textit{hazardous event} as a result of a hazard combined with a 'scenario containing conditions in which the hazard can lead to harm', although an explicit definition is omitted.

Option ii) does not restrict the \textit{hazardous event} to a point in time, by building on \textit{operational situation}, which is a 'scenario that can occur during a vehicle's life' \cite[1-3.104]{iso26262}.
The term \textit{scenario} is, however, not defined by the ISO 26262.
Indeed, the ISO 21448 introduces \textit{scenario} as a 'description of the temporal relationship between several scenes in a sequence of scenes, with goals and values within a specified situation, influenced by actions and events' \cite[3.26]{iso21448}.
But an identification of this definition with the ISO 26262 \textit{operational situation} is lacking.
Moreover, the term \emph{situation} remains undefined.

Another issue concerning both, i) and ii), is the definition of \textit{hazard}.
The ISO 26262 defines a \textit{hazard} as a 'potential source of harm caused by malfunctioning behavior of the item' \cite[1-3.75]{iso26262}, but restricts  this definition to its scope, while the general definition 'potential source of harm' coincides with the ISO/IEC Guide 51.
The ISO 21448 introduces its own definition by replacing the suggested cause of malfunctioning by '[\dots] caused by the hazardous behavior at the vehicle level'.
As to understand how the ISO 21448's and 26262's definitions of \textit{hazard} relate, comprehension of the term \textit{hazardous behavior} is essential. Even though it is frequently used within the ISO 21448, it is defined in neither standard \cite[III.A]{zhijing2022}.

Besides \textit{hazard}, other terms build on \textit{hazardous behavior}, namely \textit{triggering condition} ('specific condition of a scenario that serves as an initiator for a subsequent system reaction contributing to either a hazardous behavior or an inability to prevent or detect and mitigate a reasonably foreseeable indirect misuse' \cite[3.30]{iso21448}), \textit{performance insufficiency}  and  \textit{insufficiency of specification} ('limitation of the technical capability contributing to [\dots]' \cite[3.22]{iso21448} resp. 'specification, possibly incomplete, contributing to either a hazardous behavior or an inability to prevent or detect and mitigate a reasonably foreseeable indirect misuse when activated by one or more triggering conditions' \cite[3.12]{iso21448}).
These definitions combined with the perspective of the ISO 21448's Figures~1, 3, 4, 12 and 13 depict \textit{hazardous behavior} as a behavior possibly leading to a hazard and caused by functional insufficiencies ('insufficiency of specification or performance insufficiency' \cite[3.8]{iso21448}), activated by at least one triggering condition.
However, this cannot constitute a definition as using \textit{hazard}, \textit{triggering condition} and \textit{functional insufficiency} leads to circular referencing: their definitions already build on \textit{hazardous behavior}.

These ambiguities propagate through the definitions of the terms \textit{hazardous behavior}, \textit{scenario} and  \textit{hazardous event}. 
In particular, among others, the terms \textit{risk}, \textit{hazard} and \textit{functional insufficiency} are affected. Therefore, not even the term \textit{SOTIF} ('absence of unreasonable risk due to hazards resulting from functional insufficiencies of the intended functionality or its implementation' \cite[3.25]{iso21448}), with \textit{intended functionality} defined as 'specified functionality' \cite[3.14]{iso21448}, is clearly defined.

This also affects the applied risk classification scheme. The ISO 21448 picks up the terminology of the ISO 26262, splitting \textit{risk} in \textit{severity}, as defined previously, \textit{exposure} ('state of being in an operational situation that can be hazardous if coincident with the failure mode under analysis' \cite[1-3.48]{iso26262}) and \textit{controllability} ('ability to avoid a specified harm or damage through the timely reactions of the persons involved, possibly with support from external measures' \cite[1-3.25]{iso26262}).
The term \textit{exposure} is affected by the issues regarding the use of \emph{scenario} for the \textit{operational situation}.
The ISO 21448 does not adjust the definition of \textit{exposure}, but in the Figures~12 and 13 the term \textit{operational situation} is not included. 
Instead, \textit{exposure} is represented as a characteristic of a 'scenario containing conditions in which the hazard can lead to harm' \cite[Figures~1, 4, 12]{iso21448}.
These Figures also introduce \emph{occurrence}, a fourth factor contributing to risk.
It is represented as a characteristic of a 'scenario containing triggering conditions' \cite[Figure 12]{iso21448} which  results in \textit{hazardous behavior}. 
Its textual description reads 'probability of encountering triggering conditions during the operating phase of the functionality' \cite[6.3 Note 2]{iso21448}.
However, a definition of this term is given in neither the ISO 21448 nor the ISO 26262.

Other risk-related terms are (transitively) affected by the ambiguities of \textit{hazardous behavior}, \textit{scenario} and  \textit{hazardous event}, such as \textit{acceptance criterion} ('criterion representing the absence of an unreasonable level of risk' \cite[3.1]{iso21448}), which builds on \textit{risk}, or \textit{validation target} ('value to argue that the acceptance criterion  is met' \cite[3.33]{iso21448}) which again builds on \textit{acceptance criterion}.
Let us remark that \textit{risk} and \textit{acceptance criterion} refer solely to \textit{harm}.
However, within the ISO 21448's argumentation framework, both terms, \textit{risk} and \textit{acceptance criterion}, are also used in other contexts, where they refer to \textit{hazardous behavior} (in given scenarios) \cite[6.3]{iso21448}, the \textit{intended functionality} \cite[4.2.2]{iso21448}, \textit{hazards} (in given scenarios) \cite[3.33, 6.3, 6.5]{iso21448} or \textit{hazardous events} \cite[4.3.1, 7.4, 8.3.1]{iso21448}.
Again, the application of both terms in these contexts remains undefined by the ISO 21448.

\subsection{Critical Debate and Constructive Suggestions}
\label{subsec:sect3_critdeb_conssugg}

The prior section uncovered various deficiencies in the ISO 21448's terminology, which even lead to the ill-definedness of its central objective, the \textit{SOTIF}.
For homologation, a quantitative, empirical risk assessment is likely beneficial.
But a clear understanding of what constitutes risk is necessary for a standard-compliant quantification.
Hence, before proceeding with analyzing the quantitative validation of SOTIF we address these terminological ambiguities. 
In particular, we discuss the terms \textit{occurrence}, \textit{hazard} / \textit{hazardous behavior}, \textit{scenario}, \textit{hazardous event}, \emph{exposure} and \emph{controllability}. 
As we are solely concerned with achieving internal consistency of the ISO 21448's terminology, our suggestions may differ from other harmonized risk terminologies, e.g.\ as developed in the VVM project (\href{https://www.vvm-projekt.de/en/}{www.vvm-projekt.de/en}).
Since these aim for external harmonization, such terminologies have a broader scope.

\paragraph{Occurrence}
\textit{Occurrence} of a \textit{triggering condition}, presumably, is not defined in the ISO 21448 but contributes to the 'overall risk' \cite[Figure~12]{iso21448}.
Our proposal is to simply adopt the description in Clause~6.3 of the ISO 21448:
\begin{definition}[Occurrence of a Triggering Condition]
	Probability of encountering a triggering condition \cite[3.30]{iso21448} during the operation phase of the functionality.
\end{definition}

\paragraph{Hazard / Hazardous Behavior}
The term \textit{hazardous behavior} is not defined in the ISO 21448 even though some central terms are based on it. 
As outlined above, a definition of \textit{hazardous behavior} building on its causes or potential consequences would lead to circular referencing. 
As to solve this issue, we propose recursive definitions of the terms \textit{harm}, \textit{hazard}, \textit{hazardous behavior}, \textit{triggering condition} and \textit{functional insufficiency} with \textit{harm} as the recursion's base.
Therefore, in contrast to the ISO 21448, we suggest to use the general definition of \textit{hazard} as given in the ISO/IEC Guide 51\cite[3.2]{iso51} and noted by the ISO 26262 \cite[1-3.75 Note 1]{iso26262}, which is based solely on \textit{harm}.
Further, we propose the following definition of \textit{hazardous behavior}:
\begin{definition}[Hazardous Behavior]
	\label{def:hb}
	Behavior of the functionality which can lead to a hazard \cite[3.2]{iso51}.
\end{definition}
With these two adjustments, the entire chain, including \textit{triggering condition} and \textit{functional insufficiency}, is well defined.
Note that the ISO 21448, and hence the rest of this work, is solely concerned with 'hazards resulting from functional insufficiencies of the intended functionality' \cite[3.25]{iso21448}.

\paragraph{Scenario}
As the term \textit{scenario} is central to state-of-the-art scenario-based techniques,
its unambiguous definition is fundamental \cite[G1]{neurohr2020fundamental}.
As depicted above, the ISO 21448's definition relies on the term \textit{situation}, which is again undefined.
Notably, the original definition from Ulbrich et al.\ is not based on \textit{situation} \cite{ulbrich2015defining}. 
For resolving, we simply propose to either use the original definition verbatim or to explicitly reference the definition provided by Ulbrich et al.

\paragraph{Hazardous Event}
The main issue with \textit{hazardous event} is its origin in the ISO 26262, where it builds on terms that are not necessarily used in the same way in both standards. 
In general, we remark that such terms should be listed or annotated explicitly in Clause~3 of the ISO 21448 while referencing to the underlying terminology that is applicable.
In the context of \textit{hazardous event}, this concerns the terms \textit{hazard} and \textit{operational situation}, as outlined in \autoref{subsec:factual_analysis_terminology}.
Additionally, the term \textit{event}, which the ISO 21448 defines explicitly as 'occurrence at a point in time' \cite[3.7]{iso21448}, is contained in \textit{hazardous event}.
However, it remains unclear if a \textit{hazardous event} can be understood as a special kind of \textit{event} as implicated by the wording because its definition neither builds on the term \textit{event} nor is it explicitly restricted to a point in time.
As to achieve terminological consistency we propose to include the term \textit{event} in the definition of \textit{hazardous event}, hence restricting a \textit{hazardous event} to a point in time.
Further, we apply the definition of \textit{scenario} (ISO 21448), instead of relying on \textit{operational situation} (ISO 26262), together with the definition of \textit{hazard} (ISO/IEC Guide 51).
\begin{definition}[Hazardous Event]
	Event \cite[3.7]{iso21448} that is a combination of a hazard \cite[3.2]{iso51} and a scenario \cite[3.26]{iso21448} containing conditions in which the hazard can lead to harm \cite[3.74]{iso26262}. 
\end{definition}
Note that the definition of \textit{hazardous event} affects \textit{severity}, \textit{risk}, \textit{SOTIF}, \textit{acceptance criterion} and \textit{validation target} whose definitions are directly or indirectly based on this term.
Thus, this adaption requires annotating corresponding references.

\paragraph{Exposure}
\textit{Exposure} is affected by the same issues as \textit{hazardous event}, as it builds on both \textit{operational situation} and \textit{scenario}.
Furthermore, its definition clearly addresses functional safety by restricting \textit{hazards} to result only from failure modes.
As \textit{SOTIF} is concerned with \textit{hazards} caused by \textit{functional insufficiencies}, the ISO 21448 can not implicitly refer to the ISO 26262 for defining \textit{exposure}.
Hence, we propose a definition that refers to the correct type of \textit{hazards}:
\begin{definition}[Exposure]
	state of being in a scenario \cite[3.26]{iso21448} containing conditions in which the hazard \cite[3.2]{iso51} can lead to harm \cite[3.74]{iso26262} if coincident with the hazardous behavior (cf.~\autoref{def:hb}) under analysis.
\end{definition}

\paragraph{Controllability}
The applicable definition of \textit{controllability} from the ISO 26262 considers the reactions of the \emph{persons} involved.
It is remarked that this 'can include the driver, passengers or persons in the vicinity of the vehicle's exterior' \cite[1-3.25, Note~1]{iso26262}.
However, ADSs take over major parts of the driving and monitoring task and might even require safe functionality until a fallback operator is available.
As the ISO 21448 claims to cover all levels of driving automation, the system itself should be added as a contributor to \textit{controllability}. 

\subsection{Example}

\begin{figure*}[htb!]
	\centering
	\resizebox{0.9\linewidth}{!}{
\tikzset{>=latex}
\begin{tikzpicture}[-stealth]
	\node[draw, text width=3cm]								(1)	{\textbf{Functional Insufficiency:} Insufficient recognition of load of front vehicle};
	\node[draw, right=1.2cm of 1, circle] 					(+1) {\textbf{+}};
	\node[draw, above=1cm of +1, text width=2.2cm] 			(2) {\textbf{Triggering Condition:} Ice plates existing on truck};
	\node[draw, right=1cm of +1, text width=2.5cm]		 	(3) {\textbf{Hazardous Behavior:} Not adjusting distance};
	\node[draw, above right=-0.6cm and 1cm of 3, text width=2.5cm] 			(6) {\textbf{Hazard:} Inappropriate distance};
	\node[above left=1.2cm and 0.5cm of 6]					(anonHB1) {};
	\node[draw, right=1cm of 6, circle] 					(+2) {\textbf{+}};
	\node[draw, above=0.4cm of +2, text width=2.1cm] 		(5) {\textbf{Scenario:} Ice plates falling from truck};
	\node[draw, right=1cm of +2, text width=2.8cm] 			(4) {\textbf{Hazardous Event:} Inappropriate distance and ice plates falling from truck};
	
	\node[draw, below=1cm of 1, text width=3cm]				(10)	{\textbf{Functional Insufficiency:} Insufficient evasion maneuver for falling load};
	\node[draw, right=1.2cm of 10, circle] 					(+10) {\textbf{+}};
	\node[draw, below=1.05cm of +10, text width=2.2cm] 		(11) {\textbf{Triggering Condition:} Ice plates falling from truck};
	\node[draw, right=1cm of +10, text width=2.5cm] 		(12) {\textbf{Hazardous Behavior:} Evasive maneuver of ADS};
	\node[draw, above right=-0.2cm and 1cm of 12, text width=2.5cm] 			(13) {\textbf{Hazard:} Inappropriate evasive maneuver};
    \node[draw, below=1.8cm of 13, text width=2.5cm] 		(16) {\textbf{Hazard:} Dangerous swerving of vehicle};
	\node[below left=0.0cm and 0.4cm of 16]					(anonHB2) {};
	\node[draw, right=1cm of 13, circle] 					(+11) {\textbf{+}};
	\node[draw, above=0.4cm of +11, text width=2.1cm] 		(14) {\textbf{Scenario:} Vehicle on other lane};	
	\node[draw, right=1cm of 16, circle] 					(+16) {\textbf{+}};
	\node[draw, above=0.4cm of +16, text width=2.1cm] 		(17) {\textbf{Scenario:} Ice on road and shared roadways};	
	\node[draw, right=1cm of +11, text width=2.8cm] 		(15) {\textbf{Hazardous Event:} Inappropriate evasive maneuver and vehicle on other lane};
	\node[draw, right=1cm of +16, text width=2.8cm] 		(18) {\textbf{Hazardous Event:} Swerving with ice on road and not structurally divided roadways};

	\node[draw, right=1cm of 4, text width=2.5cm] 	(7) {\textbf{Harm:} Injuries to persons due to collision of ice plates with ADS-operated vehicle};
	\node[draw, below=0.1cm of 7, text width=2.5cm] 	(20) {\textbf{Harm:} Injuries to persons due to side collision};

	\node[draw, below=0.1cm of 20, text width=2.5cm] 	(21) {\textbf{Harm:} Injuries to ADS-operated vehicle occupants due to rolling over};
	\node[draw, below=0.1cm of 21, text width=2.5cm] 	(19) {\textbf{Harm:} Injuries to persons due to collision with oncoming traffic};
	
	\draw[->] (1) -- (+1);
	\draw[->] (2) -- (+1);
	\draw[->] (+1) -- (3);
	\draw[->] (6) -- (+2);
	\draw[->] (5) -- (+2);
	\draw[->] (+2) -- (4);
	\draw[->] (3.east) -| ([xshift=-0.4cm]6.west) -- (6.west);
	
	\draw[->] (10) -- (+10);
	\draw[->] (11) -- (+10);
	\draw[->] (+10) -- (12);
	\draw[->] (13) -- (+11);
	\draw[->] (14) -- (+11);
	\draw[->] (+11) -- (15);
	\draw[->] (12.east) -| ([xshift=-0.4cm]13.west) -- (13.west);
	
	\draw[->] (12.east) -| ([xshift=-0.4cm]16.west) -- (16.west);
	\draw[->] (16) -- (+16);
	\draw[->] (17) -- (+16);
	\draw[->] (+16) -- (18);

	\draw[->] ([yshift=0.1cm]4.east) -- ([yshift=0.1cm]7.west);
	\draw[->] ([yshift=0.2cm]15.east) -| ([xshift=-0.4cm, yshift=-0.1cm]7.west) -- ([yshift=-0.1cm]7.west);
	\draw[->] (15.east) -- (15.east-|20.west);
	\draw[->] (18.east) -- (18.east-|19.west);;
	\draw[->] ([yshift=0.2cm]18.east) -| ([xshift=-0.4cm]21.west) -- (21.west);
	
	\draw[fill=lightgray,opacity=0.35] ($(1.north west)+(-0.2,0.2)$)  rectangle ($(12.south east)+(0.2,-0.2)$);
	\node[below=0.7cm of +1] (system) {\huge\color{white}System};

	\draw[-, gray, dashed] ($(2.north west)+(-0.15,0)$) -- ($(11.south west |- 16.south west)+(-0.15,0)$);
	\draw[-] ($(11.south west |- 16.south west)+(4.0,0.5)$) node {\color{gray}\emph{Run Time}\phantom{g}};
	\draw[-] ($(11.south west |- 16.south west)+(-2,0.5)$) node {\color{gray}\emph{Design Time}};
	
	\draw[-, thick, decorate, decoration = {calligraphic brace, mirror}] ($(11.south west |- 16.south west)+(0,-0.4)$) -- node[below,yshift=-0.1cm] {Occurrence} ($(11.south east |- 16.south west)+(-0.05,-0.4)$);
	\node at (11.south east |- 16.south west) (c2) {};
	\draw[-, thick, decorate, decoration = {calligraphic brace, mirror}] ($(c2 -| 17.south west)+(0.05,-0.4)$) -- node[below,yshift=-0.1cm] {Exposure} ($(c2 -| 17.south east)+(-0.05,-0.4)$);
	\draw[-, thick, decorate, decoration = {calligraphic brace, mirror}] ($(c2 -| 18.south east)+(0.05,-0.4)$) -- node[below,yshift=-0.1cm] {Controllability} ($(c2 -| 19.south west)+(-0.05,-0.4)$);
	\draw[-, thick, decorate, decoration = {calligraphic brace, mirror}] ($(c2 -| 19.south west)+(0.05,-0.4)$) -- node[below,yshift=-0.1cm] {Severity} ($(c2 -| 19.south east)+(-0.05,-0.4)$);
	
\end{tikzpicture}}
	\caption{Example instantiating the proposed updates of the terminological risk framework of the ISO 21448.}
	\label{fig:terminology_example}
\end{figure*}
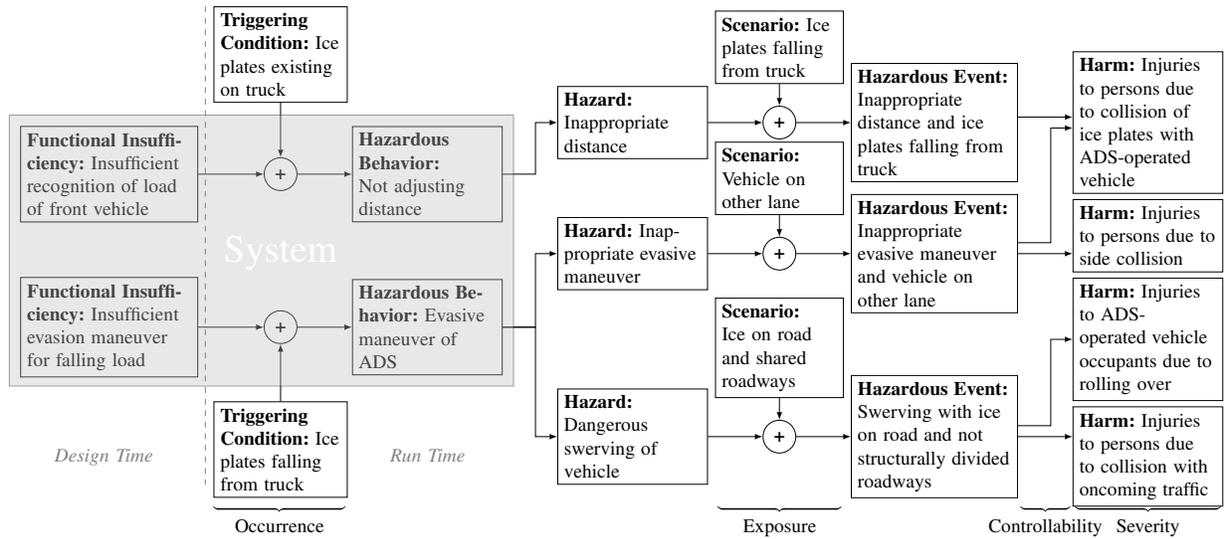

\autoref{fig:terminology_example} depicts an exemplary instance of the ISO 21448's risk model, already adapted to our terminological updates.
We focus on two functional insufficiencies introduced at design time, impacting the sense respectively plan component of the ADS: 
The first is a performance insufficiency, where the load of vehicles in front can technically only be observed restrictedly, and the latter is the specification of an insufficient evasion maneuver in case a load is falling from a vehicle in front.
The triggering conditions activating these functional insufficiencies at run time are ice plates existing on respectively falling from a truck in front of the ADS-operated vehicle.

In the first case, the resulting hazardous behavior is an unadjusted distance.
Let us emphasize that behaviors do not have to be performed actively.
In fact, omissions (such as not adjusting one's speed or distance) make up an important share of hazardous behaviors. 
This behavior directly causes the single hazard of an unadjusted distance to the vehicle in front.
The second part of the example will show how multiple hazards can be induced by one hazardous behavior.
Note that, contrary to the the ISO 21448's examples, ours does not declare collisions as hazards. 
This may be justified in certain circumstances, e.g.\ when analyzing the controllability of collisions through airbags and seatbelts.
But during a hazard analysis, specifically for behavioral safety of ADSs, one is interested in hazards not located at the end of causal chains \cite[84-3]{systemsafetyscrapbook}.
Due to this, we can identify effective controllability strategies, such as emergency braking, which can be utilized up to the point of impact.
Finally, we remark that other hazardous behaviors can lead to this exemplary hazard as well, e.g.\ not making sufficient room for a side challenger. 

Whereas an unadjusted distance may not directly be a hazard to some participant, it actualizes if the front vehicle's load starts falling, resulting in the hazardous event.
Eventually, this event can lead to a collision of the ice plates with the ADS-operated vehicle, possibly inducing harm to its passengers.

The second triggering condition is falling ice plates, and is thus equivalent to the scenario constraints of the previous example.
It can result in the ADS performing an evasive maneuver, which is not necessarily hazardous.
Although evasive maneuvers can be used to effectively mitigate hazards, such behaviors can also potentially lead to hazards, as exemplified:
\begin{enumerate}
	\item[i)] The evasive maneuver is conducted in such a way that it is a hazard to a vehicle on an adjacent lane, or
	\item[ii)] the vehicle starts to swerve dangerously, constituting a hazard to its occupants.
\end{enumerate}
Multiple harms are possible.
In the first case, we can observe injuries due to collisions with an adjacent vehicle or the ice plates (in case basic accident avoidance systems prevent later parts of the evasive maneuver to avoid collisions with the adjacent vehicle).
In the second case, swerving can result in injuries due to rolling-over or a collision with oncoming traffic.

In the example, this induces a well-defined risk decomposition: starting from the severity of the identified harms, which can be controlled after their respective hazardous event, that in turn result from an exposure to the hazard. 
This exposure is determined by the occurrences of its triggering conditions.

\subsection{Final Remarks on Terminology}

This section fixes, at least, the previously uncovered inconsistencies, thus enabling a well-founded, subsequent analysis of quantitative validation strategies suggested or required by the ISO 21448. 
Still, even if the terminological construct is internally consistent regarding risk quantification, open questions remain.
For example, external consistency to other standards can be desirable but in many cases hardly achievable. 
This concerns e.g.\ the definition of \emph{harm}, where the ISO 26262 explicitly excludes environmental and property damage, which was subsequently adopted by the ISO 21448.
Other safety standards, such as the ISO/IEC Guide 51, however, argue that the broader definition may be applicable, indicating the necessity of a rationale for the adoption from functional safety.
Moreover, an internally and even potentially externally consistent terminology does not ensure that it is actually reasonable and useful.
For instance, the \emph{intended functionality} is seen a synonym for the specified functionality, and thus the SOTIF is only concerned with the safety of the specification.
Obviously, society's demands, the engineers' intention and its resulting specification can all differ.

\section{Validation of the SOTIF using Quantitative Acceptance Criteria}
\label{sec:quantification_concept_phase}

In this section, we  analyze the framework provided by the ISO 21448 in order to examine how the SOTIF can be demonstrated when quantitative acceptance criteria are used.
Further, we discuss the derivation of validation targets from quantitative acceptance criteria based on 
a concrete example from the informative part of the ISO 21448.
We complement these discussions by some constructive suggestions, building on the adjusted terminology from the previous section.

\subsection{Quantitative Assessment of the SOTIF}

We first examine the normative parts of the ISO 21448 on quantitative assessment of the SOTIF, i.e.~Clauses 6, 7 and 9.

\paragraph{Factual Analysis}

The assessment of the SOTIF, as recommended by the ISO 21448, starts with an initial qualitative risk evaluation.
In Clause 6 of the ISO 21448 it is argued that significant similarities between ISO 26262 and ISO 21448 exist and that key terminology remains the same.
It is expressed that no ASIL classification is used in the standard, but the idea of considering severity, exposure and controllability is continued from the ISO 26262.
Instead of using ASIL, severity and controllability are treated as binary variables for the selection of SOTIF-related hazardous events, claiming that the only relevant information is whether they are zero or not.
For the estimation of severity of the harm and controllability of the hazardous event, they refer to ISO 26262-3-6 \cite{iso26262}.
Exposure is not considered in the risk evaluation.
If for a hazardous event severity or controllability are evaluated as zero, it has to be documented by sufficient evidence.
Otherwise, acceptance criteria must be formulated that will be dealt with further in Clause 7.
Lastly, certain valid rationales for the formulation of quantitative acceptance criteria are given, including GAMAB (globalement au moins aussi bon), PRB (positive risk balance), ALARP (as low as reasonably practicable) and MEM (minimal endogenous mortality).

In Clause 7, the ISO 21448 demands a systematic qualitative or quantitative analysis of potential functional insufficiencies and associated triggering conditions using expert knowledge, possibly supported by inductive, deductive or exploratory methods \cite[Table 4]{iso21448}.
Following this process, the ISO 21448 requires the evaluation of scenarios containing the identified triggering conditions, to demonstrate that the SOTIF is achievable.
This is the case if 'the residual risk of the system causing a hazardous event is shown as being lower than the acceptance criteria [\dots] and there is no known scenario that could lead to an unreasonable risk for specific road users' \cite[p. 37]{iso21448}.

In Clause 9, the ISO 21448 explicates the relevance of validation targets to argue that the acceptance criteria are fulfilled and the necessity of a strategy to provide evidence that these validation targets are met.
A suitable effort combined with an underlying rationale has to be assigned to each method-dependent validation target.
In order to reduce such validation efforts it is suggested that exposure, controllability and severity may be considered \cite[Note 4]{iso21448}. 

\paragraph{Critical Debate and Constructive Suggestions}

First, let us remark that the normative part of the ISO 21448 is rather sparse with requirements (statements using 'shall') compared to the aerospace standards \cite{arp4754a, arp4761}, the UL 4600 \cite{ul4600} or the ISO 26262. 
For each clause, 'shall'-statements are confined to the 'Objectives'-subsections which are quite abstract and unspecific. 
In particular, none of the Clauses~6, 7 or 9 \emph{require} risk to be quantified: all risks can be evaluated qualitatively. 

While it is reasonable to perform a qualitative risk evaluation in the concept phase, we question the decision to omit exposure completely  and reduce controllability and severity to binary variables. 
Firstly, we identify issues regarding the semantics of $C=0$ ('controllable in general') and $S=0$ ('no resulting harm'), since $C=0$ only requires a 'general' controllability. 
It thus allows for the probability of harm given the hazardous event to be low, but not necessarily zero, which contrasts the sharp boundary for $S=0$.
Moreover, this sharp definition leads to the (rather theoretical) issue of $S=0$ not being applicable for any identified hazard, as \emph{per definition} hazards have a non-zero probability of leading to a harm.
Defining $S=0$ as 'no harm in general' avoids this problem.

Although, intuitively, safety validation of ADSs necessitates a more rigorous risk evaluation, the suggestions of Clause~6.4 of the ISO 21448 can be viewed as a regression compared to the ASIL classification of ISO 26262.
A general reasoning why the exclusion of $E$ and a binary evaluation of $C$ and $S$ should be sufficient is missing entirely.
We deem a rationale as obligatory for such a comparatively significant change.

Another main issue is the specification of acceptance criteria of residual risk in Clause~6.5, which are a cornerstone for defining a verification and validation strategy in Clause~9. 
Again, the ISO 21448 permits qualitative and quantitative acceptance criteria, but at the same time heavily promotes quantitative acceptance criteria (GAMAB, PRB, ALARP, MEM) as examples thereof. 
However, validating quantitative acceptance criteria attached to harm, e.g.\ of the form $P(\mathit{Harm}) \le 10^{-x} / h$, immediately necessitates a quantitative evaluation of the residual risk, which again necessitates a quantitative evaluation (or, at least, estimation) of risk-related components. 

Once quantitative acceptance criteria for residual risk have been defined, it is possible to derive validation targets from them. 
As validation targets are not attached to a certain part of the risk decomposition, cf.~\cite[Definition~3.33 Note~3]{iso21448}, they can be defined on the level of harms but also for hazardous behaviors or even triggering conditions. 
As it is well known that a direct mileage-based approach for validating upper bounds on the probability of harm is not feasible for ADSs \cite{wachenfeld2016release}, validation targets that attach to a certain harm require further decomposition in order to reduce the validation effort.
A suggestion towards proper risk decomposition is given by inequalities \eqref{eq:inequality_harm} and \eqref{eq:inequality_risk}, and will be extended in future work.

\subsection{Derivation of Quantitative Validation Targets}

After discussing these normative aspects, we now consider the informative parts regarding the derivation of quantitative validation targets, namely Annexes~C.2 and, partially, C.3.

\paragraph{Factual Analysis}
The Annex C.2 of ISO 21448 provides an example for deriving a validation target from a given quantitative acceptance criterion to reduce the validation effort. In this example, the validation target is attached to a hazardous behavior and it is remarked that, according to Clause 6, every hazardous behavior is linked to an acceptance criterion.
It is assumed that, for identified incidents leading to a harm $H$, the acceptance criterion $A_H$ is given as a rate determined by 'established methods'.
Based on this, it is proposed to factorize $A_H$ as 
\begin{equation} 
	\label{eq:sotif_risk_quantification}
	A_H = R_{\mathit{HB}}\cdot P_{E|\mathit{HB}} \cdot P_{C|E} \cdot P_{S|C}\,,
\end{equation}
where $R_{\mathit{HB}}$ is the rate of hazardous behavior, and $P_{E|\mathit{HB}}$, $P_{C|E}$ and $P_{S|C}$ are conditional probabilities that are attached to the hazardous behavior, corresponding hazardous scenarios, their controllability and potential severity.
As to derive the tolerable rate of the considered hazardous behavior equation \eqref{eq:sotif_risk_quantification} is solved for $R_{\mathit{HB}}$.
The tolerable rate of hazardous behavior is then used to estimate a corresponding validation target $\tau$ describing the effort that suffices as evidence for the incident rate leading to $H$ being lower or equal to $A_H$ with confidence level $\alpha$.
Here, it is assumed that the considered hazardous behavior in a given time period follows a Poisson distribution s.t. $\tau = - \ln (1-\alpha)/R_{\mathit{HB}}$.
The example is complemented by a calculation with artificial values, where it is assumed that the acceptance criterion is a maximum frequency and $P_{E|\mathit{HB}}$, $P_{C|E}$ and $P_{S|C}$ are known from field data.

\paragraph{Critical Debate and Constructive Suggestions}
We identified three main issues in the given example that go beyond the exemplary character of the computation.

While equation \eqref{eq:sotif_risk_quantification} is not further explained, it is reminiscent of decomposing a probability into conditional probabilities.
However such a decomposition would, in contrast to the notation given, require additional conditionals for some of the probabilities e.g. $P_{S|C, E, \mathit{HB}}$ instead of $P_{S|C}$.

Moreover, there generally exists more than one hazardous behavior or potentially hazardous scenario for a harm $H$.
Thus, an aggregation over the hazardous behaviors or scenarios is required in equation \eqref{eq:sotif_risk_quantification}. 
Annex C.2.1 omits this aggregation and does not even mention it as a simplifying assumption. 

As another simplification, the quantities $P_{C|E}$, $P_{S|C}$ and $P_{E|\mathit{HB}}$ are assumed to be 'known from field data'.
In particular for ADSs, they cannot be assumed to be system-independent.
Even seemingly independent quantities such as the exposure to specific scenarios can be highly depended on systems' strategic decisions, e.g.\ its route planning.
Hence, those quantities require their own \emph{system-dependent} validation or, at least, a rationale why they can be approximated independently.

After discussing these three main issues of the suggested derivation of quantitative validation targets -- namely, incorrect use of conditionals, missing aggregation over multiple hazardous behaviors as well as assumptions on system-independence -- we now constructively propose a more general approach. It fixes the mentioned issues while picking up the idea of employing conditional probabilities.
Acceptance criteria are often specified as an upper bound of the probability of a harm in general or on the probability of a harm combined with a (minimal) severity level.\footnote{A given acceptance criterion does not necessarily have to coincide with this kind of specification. Other specifications are imaginable e.g.\ for acceptance criteria that are based on ALARP.}
As in the example of the ISO 21448, there can even be different acceptance criteria for each severity level of a given harm.
For decomposing such acceptance criteria, the causal chain proposed in \autoref{sec:relation_of_definitions} decomposes the probability of a harm by considering the hazardous events, hazardous behaviors, and triggering conditions leading to that harm.
Let us remark that many other partitions of this causal chain into events are conceivable as a basis for such a decomposition. 
For example, hazardous behaviors could be omitted entirely or a sequence of hazardous events building on each other could be included.

Let $\mathcal{H}_H$, $\mathcal{E}_E$, $\mathcal{B}_B$ and $\mathcal{T}_T$ describe the events associated with the occurrence of a harm $H$, a hazardous event $\mathit{E}$, a hazardous behavior $\mathit{B}$ and a triggering condition $\mathit{T}$, respectively.
Further, let $\mathfrak{E}$, $\mathfrak{B}$ and $\mathfrak{T}$ describe the sets of all known hazardous events, hazardous behaviors and triggering conditions that potentially lead to the considered harm $H$.
Assuming that $H$ solely occurs as consequence of an identified triple consisting of a hazardous event, a hazardous behavior and a triggering condition, 
we propose to apply the Bonferroni inequality in combination with a decomposition into conditional probabilities to derive an upper bound on the probability of $H$:
\begin{equation}
	\begin{gathered}
		P(\mathcal{H}_H)\leq \hspace{-1.8em} \sum_{E\in\mathfrak{E},B\in\mathfrak{B},T\in\mathfrak{T}} \hspace{-1.8em} P(\mathcal{T}_T) P(\mathcal{B}_B|\mathcal{T}_T)\\
		P(\mathcal{E}_E|\mathcal{B}_B,\mathcal{T}_T)P(\mathcal{H}_H|\mathcal{E}_E,\mathcal{B}_B,\mathcal{T}_T)\,.
	\end{gathered}\label{eq:inequality_harm}
\end{equation}
If $\mathcal{S}$ is a random variable encoding the different severity levels, then we can similarly derive an upper bound for the probability of harm $H$ in combination with a severity level $S$:
\begin{equation}
	\begin{gathered}
		P(\mathcal{H}_H, \mathcal{S}=S)\leq \hspace{-1.8em}\sum_{E\in\mathfrak{E},B\in\mathfrak{B},T\in\mathfrak{T}}\hspace{-1.8em}P(\mathcal{T}_T) P(\mathcal{B}_B|\mathcal{T}_T)
		P(\mathcal{E}_E|\mathcal{B}_B,\mathcal{T}_T) \\ P(\mathcal{H}_H|\mathcal{E}_E,\mathcal{B}_B,\mathcal{T}_T)
		P(\mathcal{S}=S|\mathcal{H}_H,\mathcal{E}_E,\mathcal{B}_B,\mathcal{T}_T)\,.
	\end{gathered}\label{eq:inequality_risk}
\end{equation}
The upper bounds given in these inequalities can be used for the derivation of validation targets.
We note that inequalities \eqref{eq:inequality_harm} and \eqref{eq:inequality_risk} might be too inaccurate. 
In that case, the inclusion-exclusion criterion can be applied, although this requires more validation effort as additional terms arise, cf. Gelder et al.\  \cite{de2021risk}.
In case that there exists just one tuple of hazardous event, hazardous behavior and triggering condition leading to harm $H$, the right side  of inequality \eqref{eq:inequality_risk} reduces to a term which is similar to the decomposition given in equation \eqref{eq:sotif_risk_quantification}.
However, in general, the aggregation in inequalities \eqref{eq:inequality_harm} and \eqref{eq:inequality_risk} can not be omitted, as illustrated by the example.
Thus, the estimation of validation targets requires the derivation of upper bounds for the different summands.

Further, it must be noted that in general all probabilities in inequalities \eqref{eq:inequality_harm} and \eqref{eq:inequality_risk} depend on the system, as already discussed for equation \eqref{eq:sotif_risk_quantification}.
Therefore, it is not sufficient to derive a single validation target.
Every value inserted to prove that the acceptance criterion is met requires a system-dependent validation or  at least some justification why this can be omitted.
It remains future research to investigate how the different probabilities can be obtained and validated as well as to assess which of the probabilities are good candidates for specific validation strategies, e.g.\ using simulation.

\paragraph{Example}

The exemplary instance of the adjusted risk model in \autoref{fig:terminology_example} illustrates the issues of equation \eqref{eq:sotif_risk_quantification} regarding the missing aggregation and the incorrect assumption of system-independence discussed above.
For this, \autoref{fig:terminology_example} depicts two different triples of triggering condition, hazardous behavior and hazardous event leading to the harm \textit{Injuries to persons due to collision of ice plates with ADS-operated vehicle ($H_1$)}.
Assuming that there exists a quantitative acceptance criterion for $H_1$, both triples need to be incorporated when deriving validation targets from the acceptance criterion.

Further, the hazardous events leading to $H_1$ require the existence of a truck in front respectively of a vehicle on another lane. 
The occurrence of these events depend on the operational design domain (ODD) of the ADS under investigation as well as strategic decisions of this system within its ODD.
For example, the probability of a truck in front increases if the ADS can only be deployed following another vehicle, or if its ODD is restricted to traffic jams -- even a preference for highway routes has an influence on this probability.
Thus, in this case, the exposure to the hazardous event given the hazardous behavior is not system-independent.

\section{Conclusion and Future Work}
\label{sec:conclusion}

In this work, we summarized our findings from performing a deep-dive on the ISO 21448's terminology and guidance on quantification, particularly regarding the validation strategy for driving automation at SAE levels three or above. 
The considerations have been divided into a factual analysis followed by a critical debate and constructive suggestions. 
First, as a consistent terminology is prerequisite to rigorous quantification, we provided minimally invasive suggestions to improve the terminology used by the ISO 21448. 
Second, based on these improvements, we critically reviewed the ISO 21448's guidance on quantification of the SOTIF with a special focus on acceptance criteria and the derivation of validation targets.
Finally, the suggested risk decomposition should be viewed as theoretical basis for future work.

For this, the authors aim to refine the presented risk decomposition such that the involved quantities are well-defined and estimatable in principle. 
A thorough analysis of these risk-related quantities and associated means of their estimation, through study of real-world data or simulative approaches, seems indispensable to enable a sound quantification of the SOTIF, particularly for ADSs.

\section*{Acknowledgment}
The research leading to these results is funded by the German Federal Ministry for Economic Affairs and Climate Action within the project 'VVM -- Verification \& Validation Methods for Automated Vehicles Level 4 and 5'.

\bibliographystyle{IEEEtran}
\bibliography{IEEEabrv, Literature}

\end{document}